# Enhancing IoT-Botnet Detection using Variational Auto-encoder and Cost-Sensitive Learning: A Deep Learning Approach for Imbalanced Datasets

Hassan Wasswa
School of Engineering and Information Technology
University of New South Wales
Canberra, Australia
h.wasswa@adfa.edu.au

Timothy Lynar
School of Engineering and Information Technology
University of New South Wales
Canberra, Australia
t.lynar@adfa.edu.au

Hussein Abbass
School of Engineering and Information Technology
University of New South Wales
Canberra, Australia
h.abbass@unsw.edu.au

*Abstract*—The Internet of Things (IoT) technology has rapidly gained popularity with applications widespread across a variety of industries. However, IoT devices have been recently serving as a porous layer for many malicious attacks to both personal and enterprise information systems with the most famous attacks being botnet-related attacks. The work in this study leveraged Variational Auto-encoder (VAE) and cost-sensitive learning to develop lightweight, yet effective, models for IoT-botnet detection. The aim is to enhance the detection of minority class attack traffic instances which are often missed by machine learning models. The proposed approach is evaluated on a multi-class problem setting for the detection of traffic categories on highly imbalanced datasets. The performance of two deep learning models including the standard feed forward deep neural network (DNN), and Bidirectional-LSTM (BLSTM) was evaluated and both recorded commendable results in terms of accuracy, precision, recall and F1-score for all traffic classes.

## I. INTRODUCTION

The rapid proliferation of internet of things (IoT) technology, catalyzed by its simplicity and ease of use, has anchored the technology as an integral component of our daily lives at both personal and professional levels. However, the computing components embedded in IoT devices are designed to execute dedicated tasks while often overlooking the dimensions concerned with devices' security. Moreover, aside from most IoT devices being secured using default login credentials which can easily be guessed, due to inherent limitations of storage, processing power, and battery capacity, the devices fail to support legacy security schemes such as conventional anti-virus software, and intrusion detection systems that are designed for traditional computers [1], [2].

Attackers exploit the above security vulnerabilities to launch massive, distributed attacks via IoT botnets. For example, BASHLITE and Mirai bot malware exploited over 60,000 and more than 100,000 IoT devices, respectively, between 2015 and 2017 [3]. Also, since the release of the Mirai source code in late 2017, numerous Mirai variants have been derived rendering IoT botnets the de-facto approach for malicious activities such as DDoS attacks, spam, identity theft, distributed brute force attacks, and fraud clicks [4].

To keep up with the challenge and owing to the recorded success of machine learning in fields such as image processing and computer vision [5], [6], among others, extensive research has been invested in developing effective deep learning models for detecting IoT botnet attacks. This effort has further been aided by the presence of IoT attack traffic datasets collected from various network testbeds such as N-BaIoT [1], Bot-IoT [7], CIC-IoT [8] and many more.

However, the available datasets often have at least one of two major issues: high dimensionality and class imbalance. High dimensionality leads to large weight matrices, making models heavy for the resource constrained devices. Class imbalance can result in skewed predictions, favoring the majority class. Various approaches (see Section II) such as feature selection, principal components analysis, and Auto-encoders (AEs) have been used to handle high dimensions. To address class imbalance, techniques such as random oversampling, synthetic instance generation schemes, and generative models like Generative Adversarial Network (GAN), and Variational Auto-encoders (VAE) have been used.

The above data leveling methods have shown limited effectiveness in dealing with highly imbalanced datasets like the Bot-IoT dataset where out of over 72 million instances, the "Theft" class constitutes only 1,587 (0.002% of the dataset) instances. Random under-sampling leads to significant information loss, while random oversampling leads to excessive replication of minority class instances [9], [10]. Introducing synthetic data points can lead to synthetic instances greatly outnumbering the original minority class observations, which can cause the model to only learn patterns from synthetic instances rather than real class observations [11], [12].

To enable early detection, since IoT botnet attacks begin by exploiting IoT devices before large-scale attacks, it is dire to equip these devices with lightweight defense mechanisms. This paper addresses this by proposing an approach for developing lightweight IoT botnet detection models. The focus is to improve the recognition of minority attack classes which are often missed in a highly imbalanced learning environment. The proposed approach involves utilizing a VAE to generate additional instances of the minority class(es) in a controlled manner such that synthetic samples do not outnumber the original class observations. This is because generating instances of a certain class using VAE requires prior training on instances of that class. However, minority classes usually lack sufficient samples for training a good generative model consequently leading to high reconstruction errors which accumulate with each additional synthetic instance. This justifies the need to control the number of additional instances in this work.

However, despite adding diversity to minority class instances, the above phase does not solve the class imbalance issue. Hence this work, in addition to, data augmentation utilizes cost-sensitive learning to enhance recognition of minority class instances. This is achieved by assigning a higher cost for miss-classifying a minority class instance. The approach then utilizes the classic Auto-encoder (AE) to project the high dimensional dataset to low dimensional latent space representations before passing it to the model for training and evaluation. The approach recorded plausible performance in terms of precision, recall, and F1-score, when evaluated on the Bot-IoT and CIC-IoT datasets. The main contributions of this study are:

1) We propose the idea of using VAE to generate additional minority class instances while preventing synthetic instances from outnumbering original observations. This eliminates both the possibility of learning minority class patterns from only synthetic instances and/or mistaking the few original observations as outliers.
2) By combining a VAE (for data augmentation), AE (for dimension reduction), and cost-sensitive learning (for imbalanced learning), we proposed a three-step approach for enhancing the recognition of minority class instances with a high degree of precision in a highly imbalanced dataset environment.

The rest of this paper is organized as follows. Section II presents the current state of research regarding detection of IoT botnets. This is followed by section III where a detailed description of the approach proposed in this work is presented, together with a brief description of the dataset and the various techniques used. In section VI we present and discuss the findings of this study followed by the conclusion in section VII.

## II. RELATED WORK

In this section we discuss our work in the context of: (1) dimension reduction, and (2) Data augmentation.

### A. Dimension reduction techniques deployed in prior studies

Techniques including feature selection, PCA, and most recently AEs and VAEs are the most common dimension reduction techniques. Feature selection-based techniques include filter methods and wrapper methods. Filter methods involve analysis of statistical properties such as Chi-square analysis, Correlation measure between features and the target variable, Information measure, Gini index measure, Consistency measure, and Fisher-Score analysis and have been used in studies such as [13], [14] among others to determine the optimal set of pertinent features for traffic discrimination.

With wrapper methods, a model is evaluated as features are eliminated. It has been deployed in studies including [15], [16]. On the other hand, PCA is a linear approach that works by computing Eigen values and returning Eigen vectors that point in directions of highest variance as principal components. It has been applied to intrusion detection problems such as [17], [18].

However, each of the above methods comes with its drawbacks: Filter methods produce large feature sets, lack optimal subsets, and often require manual determination of relevant features while Wrapper methods provide algorithm-dependent optimal feature sets and are computationally complex [19], [20]. PCA overlooks target variable categories, often yields sub-optimal results, needs manual determination of explained variance threshold and does not work well on data with complex nonlinear structure [21].

Due to the recent success of deep learning models, researchers have turned to Auto-encoders (AEs) to address the limitations of existing methods. By projecting the high dimensional feature vectors to low dimensional latent space vectors, AEs enable inexpensive training of classification models. Studies including [1], [22] have deployed either AE or its variants for dimensional reduction.

### B. Data augmentation techniques deployed in prior studies

Having sufficient training examples for each category is vital for training good classification models. To mitigate imbalanced distribution in training datasets, data balancing techniques such as random oversampling, SMOTE and its variants like ADASYN, have been deployed in prior studies including [18], [23].

However, the various approaches to address imbalanced datasets in machine learning have limitations. Random oversampling can increase system requirements due to the introduction of excessive redundant instances. Conventional synthetic generating methods like SMOTE and ADASYN, etc. can lead to excessive synthetic instances compared to real minority class observations and are sensitive to outliers. Training AI-based generative models like GAN and VAE requires an adequate number of training examples, which are often lacking in imbalanced scenarios for minority classes. However, AI-based models are expected to still capture some high-level characteristic patterns of minority class instances and hence can generate new instances, but with a significant reconstruction error. Training a GAN involves two models simultaneously which strain system resources. Hence, this work utilizes VAE for data augmentation, a method adopted by studies including [24], [25].

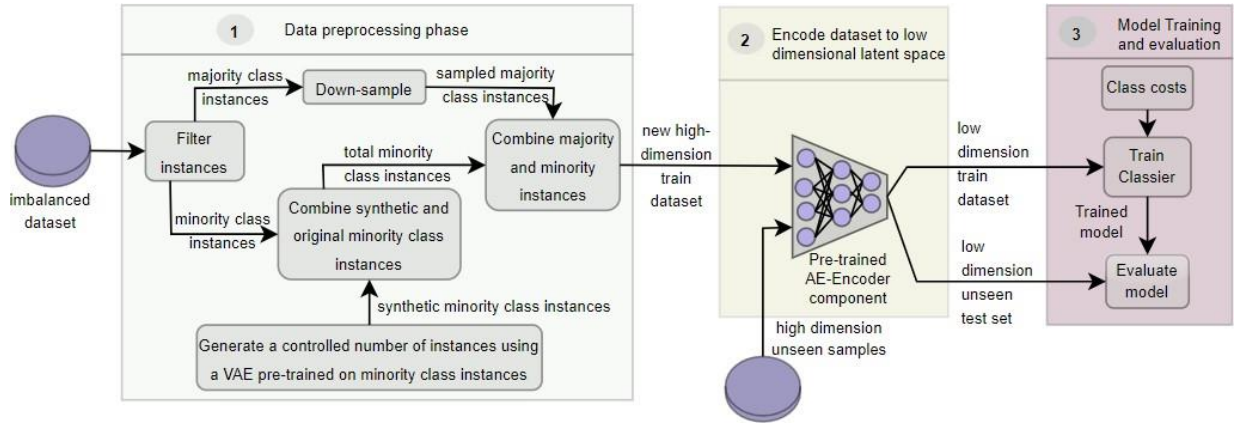

Fig. 1: Proposed VAE-Cost-Sensitive Approach for training and evaluating IoT-botnet detection models on a highly imbalanced dataset

The two fundamental components that differentiate this work from existing studies are:

1) *The nature of our approach to data augmentation*: Previous studies on data augmentation have focused on balancing dataset distributions without considering the impact of synthetic instances on classification performance. Data augmentation methods can introduce noise or estimation errors, which can accumulate and lead to the model learning patterns from synthetic instances alone. This study generates minority class data points in a controlled manner to avoid outnumbering the original minority class observations, thus preventing the model from relying solely on synthetic data points for recognition of minority class instances.

2) *Cost-sensitive learning approach to imbalanced learning*: Generating a controlled number of samples for the minority class can improve model performance in terms of minority class instance-recognition but does not completely overcome the imbalanced distribution problem. This study, in addition to data augmentation using VAE, deploys cost-sensitive learning by dynamically determining and assigning a weight to each class based on its sample size, which greatly improves the recognition of minority class instances without impacting the recognition of majority class instances in addition to keeping the systems requirements like memory minimal.

## III. PROPOSED APPROACH

The three-phase approach proposed in this study involves generating additional instances for the under-represented class(es) using a VAE, projecting the high dimension feature vectors to a low dimension latent space using AE, and then deploying cost-sensitive learning to train and evaluate learning models.

### A. Dataset

Considering that this study aimed at making minority class instances more visible to machine learning models, we chose a highly imbalanced Bot-IoT traffic dataset [26]. The published .csv full dataset constitutes over 72 million instances. Table I shows the instance distributions in terms of the five traffic categories for the full Bot-IoT dataset.

TABLE I: Full Bot-IoT Dataset distribution

| Attack Category | No. of Instances | Percentage |
|---|---|---|
| DDoS | 38,532,480 | 53.245 |
| DoS | 33,005,194 | 45.607 |
| Normal | 5,261 | 0.007 |
| Reconnaissance | 823,632 | 1.138 |
| Theft | 1,587 | 0.002 |

TABLE II: CIC-IoT Dataset distribution

| Attack Category | No. of Instances | Percentage |
|---|---|---|
| Normal | 2,616,853 | 80.870 |
| HTTP flood | 554,316 | 17.130 |
| TCP flood | 45,884 | 1.418 |
| Brute force | 12,257 | 0.379 |
| UDP flood | 6,561 | 0.203 |

The approach's efficiency was also assessed on the CIC-IoT dataset which also exhibits a significant level imbalance as shown in Table II. The dataset consists of diverse .pcap files from different IoT devices and network setups. Using the revised version of the CICFlowMeter[1] tool, 87 traffic flow features were extracted, resulting in a dataset of over 3.2 million flow records. The five multi-class labels were derived from the directory names.

### B. Data pre-processing

During this phase, the dataset was cleaned by removing invalid entries (e.g., infinity) and dropping unwanted fields (e.g., source/destination IP, packet sequence number, duplicated fields). A description of Bot-IoT dataset features can be accessed from the dataset website[2]. For this study, an experimental dataset was created by sampling 1 million instances each for "DoS" and "DDoS", 300,000 instances for "Reconnaissance", and all instances of the "Normal" and "Theft" classes for model training from the Bot-IoT dataset. Similarly, for the CIC-IoT, the "normal" class was down sampled to 1 million instances, while all instances of other attack classes were kept. These distributions were maintained for both preliminary and fully-fledged experiments. A train-test split of 80-20% was used for all model training.

[1]https://github.com/GintsEngelen/CICFlowMeter

### C. VAE for minority class instance augmentation

VAE was first introduced in [27] with the focus to mitigate the issue of latent spaces from classic Auto-encoders not being regularized. A VAE overcomes this challenge by constraining the latent space distribution so that it is as close as possible to a normal distribution. It deploys the Bayes variational inference to learn parameters of the encoder and decoder. Assuming that the dataset $D$ describes some unknown true distribution, by randomly sampling any variable, $x$ from $D$, a VAE learns an approximation of the underlying process through introduction of a latent variable, $z$ such that the joint distribution, $p_\theta(x, z)$, between $x$ and $z$ can be learned by learning some set of parameters, $\theta$. The marginal distribution $p_\theta(x)$ can hence be given as:

$$p_\theta(x) = \int p_\theta(x,z) dx \quad (1)$$

However, $p_\theta(x)$ is intractable and is therefore very expensive to compute. Hence by assuming $z$ to follow some simple distribution, the VAE assumes $p_\theta(x|z) \simeq p_\theta(z|x)$ and this way it can learn a set of parameters $\phi$ for a proxy network $q_\phi(z|x)$ that enables keeping track of $p_\theta(z|x)$. The expected value of $\log p_\theta(x)$ over $\phi$ can then be computed as:

$$E_{z \sim q_\phi(z|x)} \log p_\theta(x) = \mathcal{L}(\theta, \phi, z) + D_{KL}(q_\phi(z|x)||p_\theta(z|x)) \quad (2)$$

where $\mathcal{L}(\theta, \phi, z)$ defines the Evidence Lower Bound (ELBO), while $D_{KL}(q_\phi(z|x)||p_\theta(z|x))$ is the Kullback Leibler (KL) divergence (simply expressed as $D_{KL}$) between the approximate and true posterior distributions. VAE learns parameters, $\phi$, which are the weights of the decoder model that approximate the posterior distribution, $q_\phi(z|x)$, in a manner that maximizes ELBO hence minimizing $D_{KL}$. This way VAE allows generation of $x'$ by sampling from the latent encoding $z$ such that $x \simeq x'$.

### D. Cost-Sensitive learning

Cost-sensitive learning is a technique that concerns assigning costs or weights for wrong and right predictions, respectively, to classes in a classification problem. It is built on the core principle that the cost of miss-classifying instances is not equal for each class, especially for imbalanced distributions [28]. This way a predefined cost/weight matrix is passed to the model during the training phase. The model then learns parameters that minimize the total cost of predictions. This can be fundamental in areas where wrongly predicting instances of a particular class comes with catastrophic consequences. For example, detecting a data theft traffic instance as normal can lead to violation of data privacy and integrity since the perpetrator can get access to, and falsely modify, data maintained in the exploited information system.

[2]https://cloudstor.aarnet.edu.au/plus/apps/onlyoffice/s/umT99TnxvbpkkoE?fileId=2493653979

### E. Auto-encoder for dimension reduction

In this phase, for each dataset, we trained an AE (see Figure 1) and used the encoder components to project the original high dimensional datasets to latent space representations of 8 and 15 dimensions for Bot-IoT and CIC-IoT dataset respectively.

### F. Learning models

The work trained and evaluated models based on two architectures including a standard feed-forward Deep Neural Network (DNN) consisting of four hidden layers and Bidirectional Long Short-Term Memory (BLSTM). For BLSTM, the first hidden layer of DNN was replaced by a Bidirectional-LSTM layer. Rectified Linear unit (ReLu) was used for neuron activation except for the output layer where the softmax activation function was used. For all model training, the Adam optimizer was used while categorical-cross entropy was deployed as the loss function. We also added dropout layers to randomly drop 30% and 20% of the neuron outputs after the third and fourth hidden layers to reduce the chances of model over-fitting.

## IV. PRELIMINARY EXPERIMENTS

This phase was aimed at examining if our approach (combining data augmentation in a controlled manner with cost-sensitive learning) could improve the detection of minority class instances on a reduced dimension dataset. For this purpose, an AE with a latent space of size 8 was trained on a random sample consisting of 150,000 instances for each of DDoS, Reconnaissance and DoS attack instances and 50% and 70% of Normal and Theft class instances respectively, from the Bot-IoT dataset. At this point the dataset sample and split train-test split as described in Section III-B were utilized. The DNN model architecture described in section III-F was used for all experiments in this phase.

**Experiment 1:** The model was trained and its performance evaluated without data augmentation and cost-sensitive learning.

**Experiment 2:** To generate additional instances of the minority class, a VAE was trained on that class instances. Since each instance of the train dataset is represented as a vector, $v \in \mathbb{R}^d$, we utilized Conv1D for our VAE-convolutional layers. Two VAE models were separately trained on the "Normal" and "Theft" class instances drawn from the train set and were respectively used to generate an additional 3420 (65%) and 1,270 (80%) instances for the two classes. The model was trained on the augmented train set and its performance was evaluated on the test set.

TABLE III: Preliminary findings with Bot-IoT

| Model | Class | Experiment 1 | | | Experiment 2 | | | Experiment 3 | | |
|---|---|---|---|---|---|---|---|---|---|---|
| | | *Prc* | *Recall* | *F1* | *Prc* | *Recall* | *F1* | *Prc* | *Recall* | *F1* |
| DNN | DDoS | 1.00 | 0.99 | 0.99 | 0.98 | 0.99 | 0.98 | 1.00 | 1.00 | 1.00 |
| | DoS | 1.00 | 1.00 | 1.00 | 0.99 | 0.97 | 0.98 | 1.00 | 1.00 | 1.00 |
| | Normal | **0.25** | **0.96** | **0.40** | **0.37** | **0.93** | **0.52** | **0.75** | **0.96** | **0.84** |
| | Reconn | 0.96 | 0.96 | 0.96 | 0.99 | 0.99 | 0.99 | 1.00 | 1.00 | 1.00 |
| | Theft | **0.03** | **0.92** | **0.07** | **0.38** | **0.92** | **0.53** | **0.80** | **0.92** | **0.86** |

**Experiment 3:** To implement cost-sensitive learning, class weights were supplied to the model during training on the augmented train set. This step is the initial test of our approach. It combines dimension reduction using AE, data augmentation using VAE, and cost-sensitive learning as illustrated in Figure 1. To determine class weights for a given minority class, rigorous experiments were conducted. Beginning with some value $a \in Z^+$ and $a > 1$, as the class weight, and setting the weight of other classes to 1, the performance of the model was evaluated. If the model over-fitted to the minority class, the value was reduced by a predetermined step value $\alpha$, whereas in the case of under-fitting, the value was increased by a predetermined step value $\beta$, where $\alpha \neq \beta$, until model performance stopped improving. The class weights combination {Normal:367, DDoS:1, DoS:1, Reconnaissance:4, Theft:800} gave the best performance for our preliminary experiments.

*A. Preliminary Findings*

Model performance was evaluated based on test accuracy, precision, recall, and F-score. All three experiments produced plausible accuracy of 98.73%, 98.17%, and 99.84% for Experiment 1, 2, and 3, respectively. Table III shows the results from the three experiments in terms of precision, recall and F1-score.

Table III demonstrates that accuracy can be misleading in highly imbalanced learning scenarios, as it may not reflect the detection efficiency of minority class instances. Consequently, accuracy was excluded as a performance metric in subsequent experiments. The preliminary study revealed that our approach can potentially enhance the recognition of minority class instances. The promising results prompted the undertaking of a full study using our proposed approach as described in the subsequent sections.

## V. FULLY-FLEDGED EXPERIMENT

The preliminary experiment was extended and the performance of two additional models was evaluated on the Bot-IoT dataset. Also, the two models were evaluated on the CIC-IoT dataset. However, for the extended study, we compared the performance of the models on only two scenarios, that is, the scenario in Experiment 1 and Experiment 3 as described in the preliminary experiment. For Bot-IoT dataset, the class-weight combination obtained, and the number of additional minority classes' instances were maintained. Also, the sampled training dataset was maintained as described in Section III-B for both IoT datasets. However, parameter tuning was considered for this phase to enhance performance. On the other hand, for CIC-IoT dataset, an additional 2,450 (20%), and 1,960 (30%) data points was generated for the "brute_ force", and "udp_flood", respectively. Following the same approach described in the preliminary study, class-weights were set as
{"normal":1, "http_flood":1, "tcp_flood":2, "brute_force":5, "udp _ flood":8}. Also, unlike the Bot-IoT dataset which comprised 28 features after preprocessing, the CIC-IoT dataset
constituted 84 training features and the encoder was trained to project the dataset to a latent space of dimension 15.

*A. Experimental environment*

All experiments were conducted in python3. The used libraries include Pandas and Numpy (for data manipulation), Tensorflow-keras (for building deep learning models), and Scikit-learn (for evaluation of models).

## VI. RESULTS AND DISCUSSION

In this section the findings of this study are presented. We mainly present a per-model performance comparison of the results obtained before and after application of the proposed approach.

TABLE IV: Comparing Detection Performance with and Without the Proposed Approach on the Bot-IoT Dataset

| Model | Class | Before Our Approach | | | With Our Approach | | |
|---|---|---|---|---|---|---|---|
| | | *Prc* | *Recall* | *F1* | *Prc* | *Recall* | *F1* |
| DNN | DDoS | 1.00 | 0.99 | 1.00 | 1.00 | 1.00 | 1.00 |
| | DoS | 1.00 | 1.00 | 1.00 | 1.00 | 1.00 | 1.00 |
| | Normal | **0.64** | **0.95** | **0.77** | **1.00** | **0.95** | **0.98** |
| | Reconn | 1.00 | 0.98 | 0.99 | 1.00 | 1.00 | 1.00 |
| | Theft | **0.18** | **0.62** | **0.29** | **0.97** | **1.00** | **0.99** |
| BLSTM | DDoS | 0.99 | 0.99 | 0.99 | 1.00 | 1.00 | 1.00 |
| | DoS | 0.99 | 0.98 | 0.99 | 1.00 | 1.00 | 1.00 |
| | Normal | **0.33** | **0.95** | **0.49** | **0.99** | **0.95** | **0.97** |
| | Reconn | 0.98 | 0.98 | 0.98 | 1.00 | 1.00 | 1.00 |
| | Theft | **0.06** | **0.71** | **0.11** | **0.97** | **0.94** | **0.96** |

Tables IV and V show the test results before and after application of our proposed approach. The findings reveal that our proposed approach greatly improves the detection of under-represented class instances, in terms of precision, recall and F1-score, for both models and on both datasets. For instance, on the CIC-IoT dataset, both models produced a high precision for "tcp_flood" before the application of our approach but with very low recall implying that many "tcp_flood" instances were missed out. Our approach resolved this for both models. The results also show no significant difference in the detection performance of the two models. For traditional computers with sufficient computing resources, model size would be of less concern for a host-based intrusion detection approach. However, for IoT-botnet detection on devices with limited computing capabilities, a simpler model is more suitable. Both models in this study had four hidden layers and demonstrated acceptable performance in detecting minority class instances. Thus, either model can be deployed on resource-limited devices like IoT devices to enable early detection of IoT botnets.

TABLE V: Comparing Detection Performance with and Without the Proposed Approach on the CIC-IoT Dataset

| Model | | Before Our Approach | | | With Our Approach | | |
|---|---|---|---|---|---|---|---|
| | Class | Prc | Recall | F1 | Prc | Recall | F1 |
| DNN | brute force | 0.68 | 0.85 | 0.75 | 0.92 | 0.90 | 0.91 |
| | http flood | 0.96 | 0.99 | 0.98 | 0.96 | 1.00 | 0.98 |
| | normal | 1.00 | 1.00 | 1.00 | 1.00 | 1.00 | 1.00 |
| | tcp flood | 0.96 | 0.52 | 0.67 | 0.96 | 0.91 | 0.96 |
| | udp flood | 0.93 | 0.98 | 0.96 | 0.99 | 0.99 | 0.99 |
| BLSTM | brute force | 0.99 | 0.82 | 0.90 | 0.98 | 0.90 | 0.94 |
| | http flood | 0.96 | 1.00 | 0.98 | 0.96 | 1.00 | 0.98 |
| | normal | 1.00 | 1.00 | 1.00 | 1.00 | 1.00 | 1.00 |
| | tcp flood | 0.98 | 0.51 | 0.67 | 0.89 | 0.89 | 0.91 |
| | udp flood | 0.92 | 0.98 | 0.96 | 0.99 | 0.99 | 0.99 |

## VII. CONCLUSION

Due to the inherent resource limitations, IoT devices fail to support legacy defense schemes. This research proposed an approach for the development of lightweight, yet effective learning models feasible for deployment on the resource constrained IoT devices for early detection of IoT botnets. The approach aimed at addressing high dimensionality and imbalanced distribution exhibited by IoT attack datasets, which significantly enhanced the recognition of minority class instances. For instance, after its application, the BLSTM model, trained on the Bot-IoT dataset, significantly improved the recognition of "Normal" and "Theft" classes in terms of all metrics.